\documentclass[a4paper,conference]{IEEEtran}
\usepackage{graphicx}
\usepackage{comment}
\usepackage{amsmath,amssymb} 
\usepackage{color}
\usepackage{subcaption}

\newcommand{\matr}[1]{\mathbf{#1}}

\ifCLASSINFOpdf
\else
\fi

\begin{document}

\title{FatNet: A Feature-attentive Network \\ for 3D Point Cloud Processing}

\author{\IEEEauthorblockN{Chaitanya Kaul}
\IEEEauthorblockA{Department of Computer Science\\
University of York\\
Heslington, York, UK, YO10 5DD\\
Email: ck807@york.ac.uk}
\and
\IEEEauthorblockN{Nick Pears}
\IEEEauthorblockA{Department of Computer Science\\
University of York\\
Heslington, York, UK, YO10 5DD\\
Email: nick.pears@york.ac.uk}
\and
\IEEEauthorblockN{Suresh Manandhar}
\IEEEauthorblockA{NAAMII \\
Katunje, Kathmandu, Nepal\\
Email: suresh.manandhar@naamii.org.np}}

\maketitle

\begin{abstract}
The application of deep learning to 3D point clouds is challenging due to its lack of order. Inspired by the point embeddings of PointNet and the edge embeddings of DGCNNs, we propose three improvements to the task of point cloud analysis. First, we introduce a novel feature-attentive neural network layer, a FAT layer, that combines both global point-based features and local edge-based features in order to generate better embeddings. Second, we find that applying the same attention mechanism across two different forms of feature map aggregation, max pooling and average pooling, gives better performance than either alone. Third, we observe that residual feature reuse in this setting propagates information more effectively between the layers, and makes the network easier to train. Our architecture achieves state-of-the-art results on the task of point cloud classification, as demonstrated on the ModelNet40 dataset, and an extremely competitive performance on the ShapeNet part segmentation challenge.
\end{abstract}

\IEEEpeerreviewmaketitle

\section{Introduction}

Deep learning on 3D point clouds has progressed at a fast rate since the introduction of PointNet \cite{pointnet}. Due to the lack of any arrangement of a set of $N$ points lying in a $D$-dimensional space, the network needs to be invariant to their reorderings. The first papers to point this out as a potential research problem were PointNet \cite{pointnet} and Deep Sets \cite{deepsets}. They use permutation invariant and permutation equivariant functions respectively to process the points to map the data to a symmetric function, which results in the desired representation of the data. Further research \cite{pointnet2} has taken the idea of PointNet forward and applied it to various domains. Locality information has also been added to the architecture in Dynamic Graph CNNs (DGCNNs) \cite{dgcnn}, where networks look at a local region in space rather than the entire point cloud in terms of its global coordinates.

Inspired by both the PoinNet and DGCNN architectures, we address the notion of simultaneously looking both globally \emph{and} locally at the input point clouds to extract meaningful and appropriately weighted features from them. We learn meaningful global point cloud embeddings using the shared Multi Layer Perceptron (MLP) to obtain a symmetric function over the entire point cloud, making it invariant to any permutations in the input. We combine this representation with the dynamic locality information from DGCNNs \cite{dgcnn} to create a \emph{layerwise} representation for the input points that contains both global, as well as local information. We enhance the architecture further by rescaling each embedding produced by our layer, via squeeze-excite \cite{se}, such that the most important point representations are always used by the network to perform the task at hand. Furthermore, we apply attention over different feature aggregations and we use residual identity mappings \cite{Resnet} to propagate information between embeddings. These connections also lead to better gradient backpropagation through the network and this leads to stable learning. In summary, our contributions are:
\begin{enumerate}
  \item We propose a novel attention-infused layer, a FAT layer, for 3D point cloud processing that optimally combines both global point-based and local edge-based embeddings, via non-linearly derived weightings.
  \item We apply weightings over two different feature aggregation methods that is better than either aggregation alone.
  \item For the first time, network learning for processing 3D point clouds is enhanced with residual connections, with upscaling of embedding dimension handled by shared-weight MLPs.
  \item An extensive evaluation shows state-of-the-art results on the ModelNet40 classification task, highly competitive results on the ShapeNet parts segmentation task, and exceptional robustness to random input point dropout. Ablation studies confirm the effectiveness of the constituent components of our network.
\end{enumerate}
The rest of the paper is organized as follows. In Sect.~\ref{literature}, we present the background literature. Section \ref{ch5:arch} describes the FatNet  architecture both in outline and detail. We follow that by first an evaluation of 3D point cloud classification (Sect.~\ref{ch5:eval:classify}) and then an evaluation of 3D point cloud segmentation (Sect.~\ref{ch5:eval:segment}), both using standard benchmark datasets and compared to state-of-the-art networks.
The evaluation sections also detail further experiments, such as robustness to missing input points, and ablation studies that show the effective contribution of each of our network components. Finally, we present our conclusions 
in Sect.~\ref{ch5:summary}.

\section{Related Work}
\label{literature}

\begin{figure*}[ht]
\begin{center}
\includegraphics[width=1.0\linewidth]{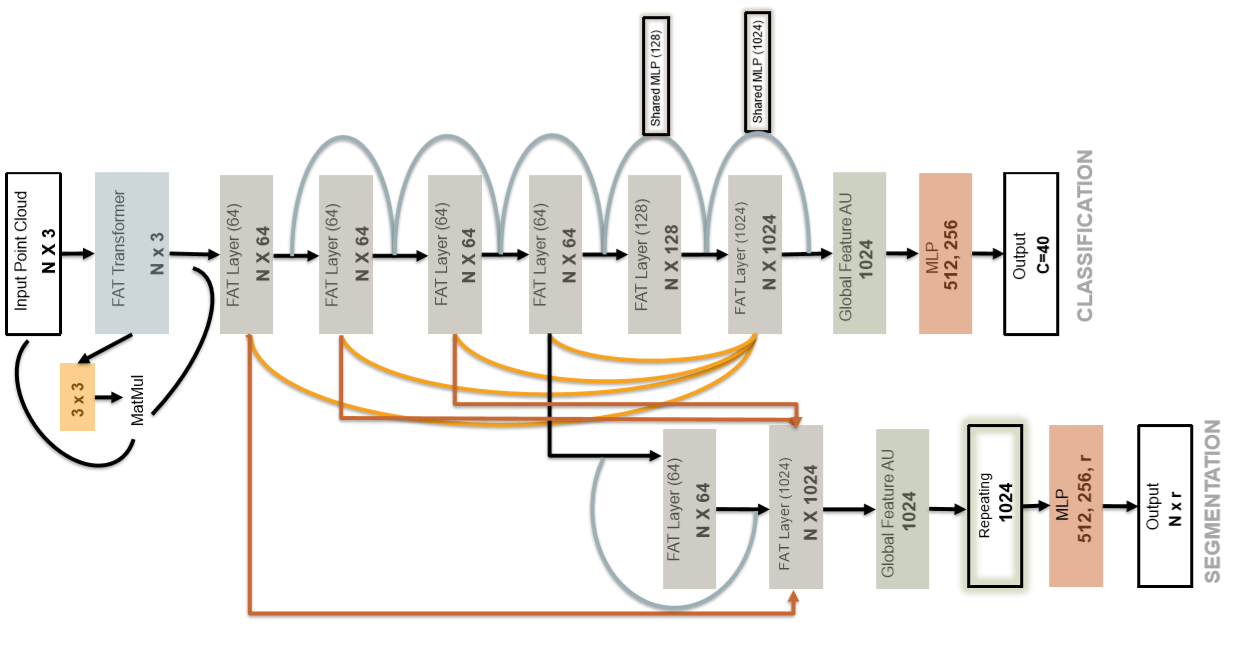}
\end{center}
\caption{The proposed FatNet  architecture for the classification and segmentation tasks. Upper path: classification over $c$ classes. Lower path: segmentation over the $r$ parts of class $c$.}
\label{final_arch}
\end{figure*}

\subsection{Deep Learning on 3D Point Clouds} For 3D points, there is no grid-like structure to apply the convolution operation on. So several earlier techniques took the 3D points and projected them on a 2D grid like structure, following which, 2D CNNs were used on the projections. Networks such as MVCNN \cite{mvcnn} sample 80 views from the point clouds and project them to 2D, which are then processed by a CNN. This is followed by a view-pooling operation, the output of which was fed into another CNN which is used for classification. Another grid-like representation is the voxel based one. VoxNet \cite{voxnet} sample 3D points into a $32 \times 32 \times 32$ 3D occupancy grid which is processed by a 3D CNN. The networks that require projections have some notable drawbacks, the most important of which is the computational expense. Storage and processing of multiple views or voxel based data is computationally expensive and projection of the data on to a view point or a small 3D occupancy grid leads to the loss of important information.
The first works in processing the point clouds directly were based on the observation that the network layers need to learn an order-invariant representation of the input points. Two approaches were introduced in this regard, namely permutation invariance and permutation equivariance. The permutation invariance approach was introduced in the PointNet \cite{pointnet} paper, where the authors hypothesised learning a symmetric function to account for all $M$! permutations of the input points of a point cloud. The major drawback of this approach was that the network only looked at the global point coordinates so this representation was not as robust. Locality information was introduced by the authors in PointNet++ \cite{pointnet2} where they used farthest point sampling to uniformly sample points to build a local region, and grouped them based on a radius-based ball query to define a local region. The PointNet architecture was the applied to this local space. These networks have found applications in exploring spatial context for scene segmentation \cite{spatialcontext}, 3D object detection from RGB-D data \cite{frustrum}, and even normal and curvature estimation \cite{pcpnet} from noisy point clouds. Networks such as SO-Net \cite{sonet} perform hierarchical feature extraction using Self-Organising Maps (SOMs). The networks discussed so far capture local geometric information using local points. Dynamic Graph Convolutional Neural Networks (DGCNNs) \cite{dgcnn} compute edge features based on these points to learn a relationship between the points and its features. These networks recompute the graph based on the nearest neighbours at every layer, and hence they are termed as dynamic. Networks such as SpiderCNN \cite{spidercnn} (SpiderConv) and PointCNN \cite{pointcnn} (X-Conv) define convolution operations on point clouds based on local geometric information, rather than using a symmetric function to process them. 

\subsection{Attention mechanisms} 
Self attention mechanisms aim to learn context beyond a networks receptive field. The first successful work in incorporating such a mechanism in CNNs was squeeze and excitation networks \cite{se}. They proposed to global average pool feature map information into a single vector creating a global representation, that was then encoded-decoded and passed through a sigmoid gating to generate attention weights for each feature map in an activation output. The maps were then scaled via multiplication with these attention weights. SE Blocks have been extensively used in object detection \cite{seobjectdetection}, image segmentation \cite{sesegmentation} and scene classification \cite{sesceneclass} to name a few applications. We extensively use variations of SE Blocks as the attention mechanisms for our feature-aware layers and global aggregation units in FatNet. 

\section{Network Architecture}
\label{ch5:arch}

In this section, we first provide an overview of the FatNet architecture, we then describe the new feature-attentive layer. For clarity, we break this down into a purely textual description, followed by a  more formal description of the layer's processes.  Finally we describe our new feature aggregation module that weights max pooling and average pooling.

\subsection{Overview}

The proposed FatNet  architecture is shown in Fig.~\ref{final_arch}. It takes a $N \times 3$ point cloud as input and outputs a class label or a set of per-point semantic segmentation labels. The input is fed into a transformer net that employs FatNet layers (without attention) to regress a $3 \times 3$ transform. This aligns the point cloud, which is then fed into a series of FatNet layers (with attention) to create a permutation-invariant embedding of the points. 

We use residual connections to transfer information between layers, allowing us to train deeper and more stable network architectures. For layers where the dimension of the embedding is increased, we cannot employ the usual identity mapping to pass the input to the output in the context of residual blocks. To deal with this we employ a shared-weight MLP, before the addition of the embedding from layer $l-1$ to the $l^{th}$ layer embedding. Two such MLPs are shown in the upper part of Fig.~\ref{final_arch}.

To exploit feature reuse and obtain a better global contextual aggregation, we concatenate the output of every previous FatNet layer (FAT layer) before feeding this feature set forward into the final 1024-D FAT layer (see orange connections in Fig.~\ref{final_arch}). The output of this final 1024-dimensional embedding is then aggregated using our novel FatNet  aggregation unit (Sect. \ref{globalagg}), to get a 1024 dimensional vector representing the input point cloud. This vector is used to classify or segment the point cloud using a generic MLP. For the segmentation tasks, the output of the MLP is reshaped to get a per-point semantic segmentation.

\subsection{FAT Transformer Net}
\label{spatialtransformer}

The FAT transformer network contains 3 FAT Layers (without attention) which embed the point cloud into a 64, 128 and a 1024 dimensional space respectively. The output embedding from the final layer is aggregated into a global context (pooled) feature vector that is fed into a multi-layer perceptron with two hidden layers of size 512 and 256. The output is a final dense layer of size 9 that is reshaped into a $3 \times 3$ transformation matrix, the elements of which are the learnt transformation values for the point cloud. A simple matrix multiplication of this matrix with the input point cloud, aligns the points in 3D space. The architecture can be seen as a mini version of the FatNet architecture. FatNet without any alignment of input point clouds on the ModelNet40 dataset performs inferior to even the most basic transformation matrix that 1D convolutions can learn.

We compared the results of our transformer network to the one presented in PointNet. We observe a 0.6\% performance gain over processing point clouds unaligned, and a 0.3\% performance gain over using the PointNet transformation network using our approach. Our results are summarized in Table \ref{tab::tnet}.

\begin{table}[h]
\begin{center}
\begin{tabular}{p{6.0cm} p{1.0cm}}
\hline
Transformation & Acc (\%) \\
\hline\hline
FatNet + Unaligned ModelNet40 (No Transformer) & 92.6\\
FatNet + T-Net (PointNet Input Transformer) & 92.9\\
FatNet + FAT T-Net (Our Contribution) & 93.2\\
\hline
\end{tabular}
\end{center}
\caption{Effect of input transformer on the FatNet output.}
\label{tab::tnet}
\end{table}

\subsection{The Feature-attentive Layer (FAT layer)}
\label{sawlayer}

\begin{figure}[htb]
\begin{center}
\includegraphics[width=0.95\linewidth, scale=0.95]{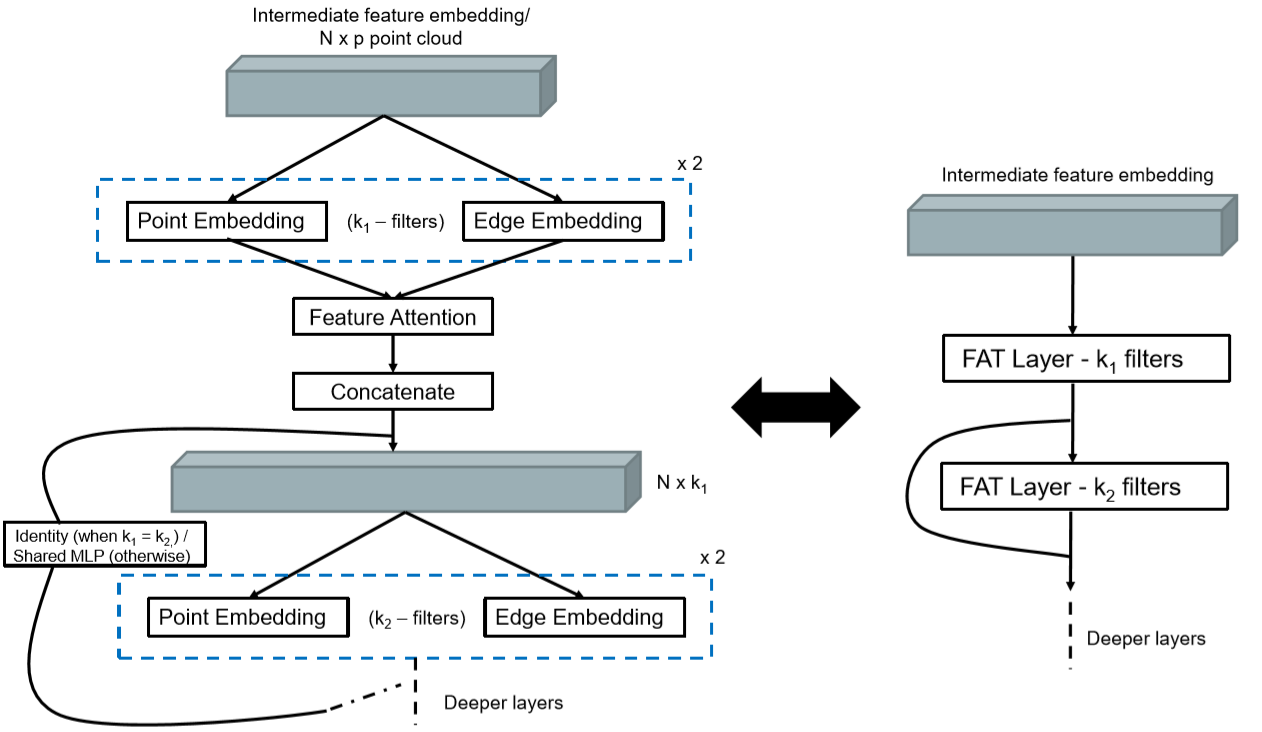}
\end{center}
\caption{The layer structure employed in FatNet.}
\label{singlelayer}
\end{figure}

The building blocks of a FAT layer can be seen in Fig.~\ref{singlelayer}. Information in the layer propagates to the next layer via skip connections that can adjust the dimension of the previous layer's embedding via shared MLP structures, if necessary. Each FAT  layer computes separate point and edge embeddings that are rescaled using a squeeze-excite feature attention mechanism. We concatenate the rescaled point and edge features and pass them to the next layer. 

The point embedding operation corresponds to a PointNet-style feature extraction using a shared-weight MLP (conv1D), while the edge embedding is achieved via dynamic graphs, as in the DGCNN architecture. 
We use a squeeze-excite feature attention block, depicted in Fig.~\ref{feature_attention_block}, to calculate attention weights for each individual feature embeddings. In the feature embedding block, we learn the attention weights for each point/edge feature using a shared-weight encoder-decoder with two dense layers. The input embeddings are first maxpooled to get a single vector, after which they are fed into the encoder-decoder. The first dense layer takes an input the size of the maxpooled output. The second dense layer is the bottleneck layer that compresses this input via some compression ratio (a ratio of eight is used). The bottleneck representation is then upsampled back to the original dimension and passed through a sigmoid gating to get the attention weights. These weights scale the individual point/edge features to recalibrate their effect on the output. 

The weighted output from the edge embedding layer is then (re)maxpooled across the neighbors, concatenated with the output from the point embedding MLP, and fed into the next layer. Each point embedding computes an embedding using conv1D, batch normalization (BN) and LeakyReLU as a part of its output estimation. The output from the second point embedding in the FAT layer propagates via a residual connection to the output of the first point embedding of the next layer. The edge embedding has a similar structure of conv1D - BN - LeakyReLU, which are used to learn edge weights. 

\subsection{Formal definition of FAT layer processes}

\subsubsection{Notation}
For the $l$th FAT  layer, we denote the $D$-dimensional embedding of the set of $N$-points as $\matr{X}^l = \{ \mathbf{x}^l_1, \dots , \mathbf{x}^l_N \} $, where $\mathbf{x}^l_i \in \mathbb{R}^D$. Furthermore, we define the set of $K$ nearest neighbour (kNN) edges for the $i$th point ($i \in \{ 1 \dots N \}$) in the $l$th layer to be 
$\Delta\matr{X}^l_i = \{ \mathbf{x}^l_{j_1} - \mathbf{x}^l_{i}, \dots , \mathbf{x}^l_{j_K} - \mathbf{x}^l_{i} \} $, where $j_k: j_k \in \{ 1, \dots , N \} \wedge (j_k \ne i) $. 
We denote MLP-based point embedding and edge embedding functions of the $l$th layer by $h^l_j(.)$ and $e^l_j(.)$ respectively, where $j \in \{1,2\}$ is used to represent the fact that there are a pair of concatenated shared-weight MLPs within each of the point and edge embeddings, as shown in Fig.~\ref{singlelayer}.

\subsubsection{Point embedding}
The first point embedding function in the $l^{th}$ layer, $h^l_1(.)$, applies an embedding to the set of points from the layer $l-1$:
\begin{equation}
h_1^l(\matr{X}^{l-1}) = \{h^l_1(\mathbf{x}^{l-1}_1),...,h^{l}_1(\mathbf{x}^{l-1}_N)\}, \ h^l_1: \mathbb{R}^{D^{l-1}} \to \mathbb{R}^{D^l} 
\end{equation}
where $D^{l}$ is the dimension of the $l^{th}$  layer embedding and $h^l_1$ is implemented using a shared-weight MLP. The output of this first shared MLP layer is $\mathbf{S}^l_1 = \delta(B^l_1(h_1^l(\matr{X}^{l-1})))$, where $\delta$ is the LeakyReLU activation function and $B^l_j$ is the $j^{th}$ ($j \in \{1,2\}$) batch normalization (BN) in the $l^{th}$  layer. This output is then fed into another shared MLP/BN combination. The output can be represented as, $\mathbf{S}^l_2 = B^l_2(h^l_2(\mathbf{S}^l_1))$.

\subsubsection{Edge embedding}
We determine the $k$ nearest neighbours ($k=20$) in the feature space at some layer, 
Then one pass of the input $\Delta\matr{X}^{l-1}_i$ through conv1D-BN-LeakyReLU gives an initial edge embedding $\mathbf{E}^l_1 = \delta(B^l_1(e^l_1(\Delta\matr{X}^{l-1}_i)))$. 
Applying another conv1D-BN gives $\mathbf{E}^l_2 = B^l_2(e^l_2(\mathbf{E}_1))$. 

\subsubsection{Feature attention}
We employ a shared-weight encoder-decoder structure to be applied to both point and edge embeddings. To ensure edge embeddings are the same size as point embeddings for this structure, they are max pooled over all $K$ nearest neighbours. We denote the shared-weight encoder-decoder encoding weights as $\mathbf{W}^l_1$ and the decoding weights as $\mathbf{W}^l_2$. Thus the point attention weight, $\mathbf{S}^l_w$, and edge attention weights, $\mathbf{E}^l_w$, are computed in the following way:
\begin{equation}
\mathbf{S}^l_w = \sigma(\mathbf{W}^l_2, \delta(\mathbf{W}^l_1, \mathbf{S}^l_2)), ~~~\mathbf{E}^l_w = \sigma(\mathbf{W}^l_2, \delta(\mathbf{W}^l_1, \mathbf{E}^l_2)  
\end{equation}
where $\sigma(.)$ represents the sigmoid function.
These then scale the embeddings as:
\begin{equation}
\mathbf{S}^l_2 \leftarrow \mathbf{S}^l_2 \cdot \mathbf{S}^l_w, ~~~ \mathbf{E}^l_2 \leftarrow \mathbf{E}^l_2 \cdot \mathbf{E}^l_w,
\end{equation}
The weighted output of the edge embedding is then maxpooled over its nearest neighbours, again to match its dimensions with the point embedded output. These values are now the same in dimension and hence can be concatenated together to form the output of the FAT layer, $\matr{X}^l = [\mathbf{S}^l_2:\mathbf{E}^l_2]$, where $[\ :\ ]$ denotes concatenation. This point embedding is fed into the subsequent FAT layers.

\begin{figure}[t]
\begin{center}
\includegraphics[width=0.95\linewidth,scale=0.95]{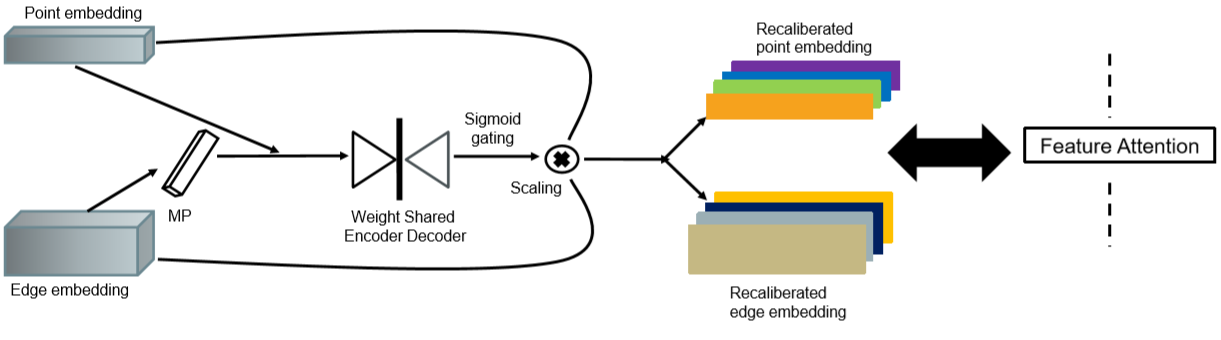}
\end{center}
\caption{The feature attention block present in every FAT layer.}
\label{feature_attention_block}
\end{figure}

\subsection{Global Feature Aggregation (GFA) Block}
\label{globalaggunit}

\begin{figure}[h]
\begin{center}
\includegraphics[width=0.95\linewidth,scale=0.95]{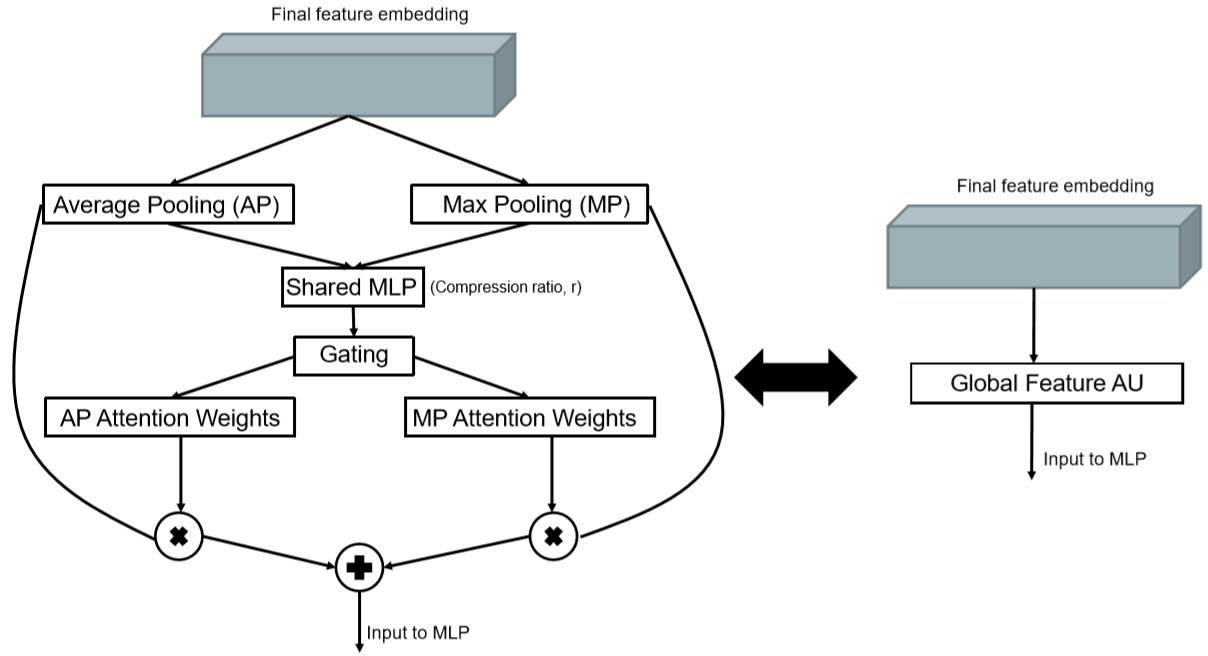}
\end{center}
\caption{Average pooling and max pooling operations are passed through a shared-weight encoder-decoder (compression ratio = 16) followed by a sigmoid gating to give attention weights for both poolings. The scaled outputs are added together to give the our Global Feature Aggregation (GFA) block.}
\label{global_feature_aggregation}
\end{figure}

Existing point cloud processing architectures embed the points into a higher dimensional feature space and follow up with a max pooling global feature aggregation. Given that all the information present in the embeddings is combined into this global aggregation, it certainly demands further study to see if better encodings can be created of the information extracted via the conv1D operations. To the best of our knowledge, no prior work has been done to study or modify this global feature aggregation since the inception of PointNet itself. One of the contributions of PointNet showed that max pooling aggregation is superior to average pooling. However, that by no means suggests that average pooled features cannot contribute to the overall network performance. 

To this end, we propose to take the average pooled and max pooled features and learn the attention weights to scale their individual influences on the output. We take the final feature embedding from the 1024-D FAT layer and simultaneously max pool and average pool them to obtain two vectored representations. These representations are both fed into a shared weighted encoder-decoder similar to the one in Fig.~\ref{feature_attention_block} with a compression ratio of 16, the output of which is passed through a sigmoid gating to get the weights for both aggregations. These weights then scale the outputs of their respective pooling operations and are then added together. In training, the network learns to appropriately weight aggregated features from both feature aggregation techniques. Later we demonstrate that this gives better results than max pooling alone. Our Global Feature Aggregation (GFA) block is illustrated in Fig.~\ref{global_feature_aggregation}.

\section{Evaluation of Point Cloud Classification}
\label{ch5:eval:classify}

We use the ModelNet40 dataset \cite{shapenet} for classification. ModelNet40 contains 12,311 CAD models for 40 objects in 3D. The models are man made for the 40 object categories and the dataset is divided into a training and test split of 9,843 and 2,468 respectively. We use the same split for our experiments and report the results on the test set.

To demonstrate that our method improves performance over PointNet and DGCNN, we use the same pre-processing as in their experiments. Hence, for each of the 3D models, we sample 1024 points uniformly from the mesh faces and normalise these points to a unit sphere. The dataset contains points as well as surface normals. We only consider the point cloud coordinates and discard the remaining information for our experiments. During training, we use data augmentation to add variations in our dataset. Random rotations and scaling are added along with per point jitter to perturb the location of the points. This is the same strategy as was used for data augmentation in PointNet++ \cite{pointnet2} training. 

We use Tensorflow \cite{tensorflow} for all our implementations, unless explicitly stated otherwise. For the classification task, we use the same training setting as in PointNet \cite{pointnet}. Our classification network was trained using a batch size of 8, on one Nvidia GTX 1080 Ti and categorical cross entropy loss. We use Adam \cite{adam} with a learning rate of 0.001, which is reduced using a decay parameter of 0.7. The decay rate for batch normalization is 0.9. We train our model for 250 epochs. 

\begin{table}
\begin{center}
\begin{tabular}{p{3.0cm} p{1.8cm} p{1.8cm}}
\hline
Method & Class Accuracy (\%) & Instance Accuracy (\%)\\
\hline\hline
3D ShapeNets \cite{shapenet} & 77.3 & 84.7\\
VoxNet \cite{voxnet} & 83.0 & 85.9\\
Subvolume \cite{subvolume} & 86.0 & 89.2\\
ECC \cite{ECC} & 83.2 & 87.4\\
PointNet \cite{pointnet} & 86.0 & 89.2\\
PointNet++ \cite{pointnet2} & - & 90.7\\
KD-Net (Depth 10) \cite{kdnet} & 86.3 & 90.6\\
KD-Net (Depth 15) \cite{kdnet} & 88.5 & 91.8\\
DGCNN \cite{dgcnn} & 90.2 & 92.2\\
SO-Net \cite{sonet} ($2048 \times 3$)& 87.3 & 90.9\\
SpiderCNN \cite{spidercnn} & - & 90.5\\
PCNN \cite{pcnn}  & - & 92.3 \\
PointCNN \cite{pointcnn} & 88.1 & 92.2 \\
\hline
FatNet Vanilla (Ours) & 90.0 & 91.8\\
FatNet (Ours) & \textbf{90.6} & \textbf{93.2}\\
\hline
\end{tabular}
\end{center}
\caption{Classification results on the ModelNet40 dataset.}
\label{ModelNet40}
\end{table}
Our classification results are summarised in Table \ref{ModelNet40} and compared with other published architectures. All networks shown are trained on 1024 points unless stated. Our network achieves state of the art results in terms of both class, as well as instance accuracy. As our architecture was inspired by both PointNet and DGCNN, it is useful to see if our methodology provides any improvement over the two. We implement FatNet vanilla, which does not use any attention mechanisms in the FAT layers and aggregates global features using max pooling. FatNet vanilla outperforms PointNet's class and instance accuracy by 3.0\% and 2.6\% respectively, which can be seen as a considerable performance gain. Re-running DGCNN from the author's official github repository gave a class and instance accuracy of 89.2\% and 91.6\% respectively. As we used that implementation as our backbone, we compare with the results we obtained during our runs. We observe a 0.8\% and 0.2\% performance gain over the two accuracy metrics by just combining the two embeddings, which we believe emperically justifies the need for FAT layers. FatNet itself, on the other hand, considerably outperforms every architecture that processes points in their raw form. We outperform DGCNN, the previous state of the art, by 0.4\%  in terms of class accuracy and by 1\% in terms of instance accuracy. The best in class instance accuracy was previously obtained by PCNN \cite{pcnn}. Our architecture outperforms their results by 0.9\% with approximately 63\% fewer parameters.

\subsection{Model Performance}

\begin{table}[h]
\begin{center}
\begin{tabular}{p{2.0cm}|| p{1.2cm} p{1.2cm} p{1.2cm} p{1.2cm}}
\hline
Method & Parameters ($\times10^8$) & Model Size (MB) & Inference time (ms)  & Accuracy (\%) \\
\hline\hline
PointNet \cite{pointnet} & 3.5 & 40 & 16.6 & 89.2\\
PointNet++ \cite{pointnet2} & 1.5 & 12 & 163.2 & 90.7\\
DGCNN \cite{dgcnn} & 1.9 & 21 & 27.2 & 92.2\\
PCNN \cite{pcnn} & 8.2 & 94 & 117.0 & 92.3\\
FatNet & 2.7 & 41 & 52.3 & 93.2\\
\hline
\end{tabular}
\end{center}
\caption{Model complexity vs performance for different architectures. Our model provides a good trade-off between model complexity and accuracy (as reported on the ModelNet40 dataset). Our forward pass for inference is considerably faster than PointNet++ and PCNN approaches along with a manageable model size.}
\label{perfsawnet}
\end{table}

Table \ref{perfsawnet} shows the overall model performance of FatNet  with respect to its model complexity. We only compare our model with officially reported values of the architectures. FatNet  contains 2.7 million parameters making it fairly lightweight. It has fewer parameters than PointNet even though it combines the type of embeddings seen in both PointNet and DGCNN. This is due to the fact that feature transformers used in PointNet consume a lot of parameters. Our inference time is understandably slightly slower than that of PointNet and DGCNN, but it is over 3 times faster than PointNet++ and over 2 times faster PCNN, while also being more accurate than the two. Overall, our network achieves a good trade-off between model complexity and performance. It is useful to know that the inference times depend on the type of GPUs used. We used Nivida GTX 1080Tis for all our experiments which have lesser number of cores, compared to the Nvidia Titan X which was used by the respective authors to run DGCNN, PointNet, PointNet++ and PCNN.

\subsection{Further Experiments on ModelNet40}
\label{ch5:eval:furtherclassify}

We used the ModelNet40 dataset for a series of experiments to finalise our network architecture, which we now discusses in detail. We conducted several experiments initially using the PointNet architecture to observe the effects of different point embedding styles as well as the effect of nearest neighbours on edge embeddings computed for our feature-attentive layer. The results from the experiments showed an increase in performance when skip connections were added to PointNet. Also, using 20 nearest neighbours to compute edge embeddings showed the best trade-off between model complexity and accuracy.

\subsubsection{Experimenting with FAT layers}
\label{spatialawareness}

\begin{figure}[h]
\centering
\begin{subfigure}[b]{0.45\textwidth}
  \centering
  \includegraphics[width=0.95\linewidth]{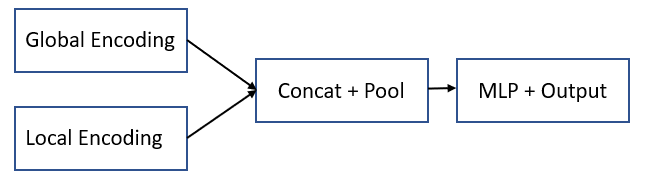}
  \caption{}
  \label{ablation1}
\end{subfigure}

\begin{subfigure}[b]{0.45\textwidth}
  \centering
  \includegraphics[width=0.95\linewidth]{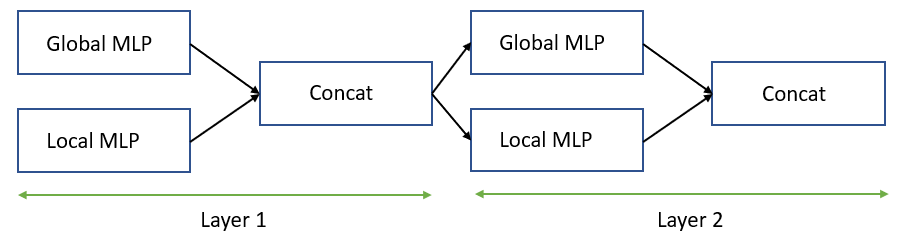}
  \caption{}
  \label{ablation2}
\end{subfigure}

\begin{subfigure}[b]{0.45\textwidth}
  \centering
  \includegraphics[width=0.95\linewidth]{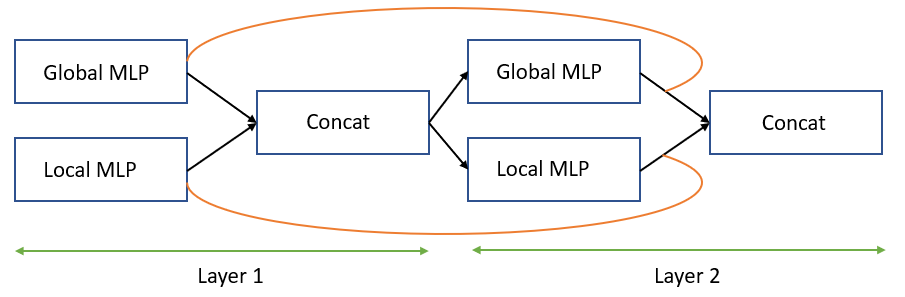}
  \caption{}
  \label{ablation3}
\end{subfigure}
\caption{Different ablation setups for the FAT layer.}
\label{spatialablations}
\end{figure}

\begin{table}[h]
\begin{center}
\begin{tabular}{l c}
\hline
Method & Accuracy (\%) \\
\hline\hline
Combine at end & 88.9\\
Combine per layer & 89.6\\
Residual + Combine per layer & 90.0\\
Residual + Combine per layer - 1 FC & 63.6\\
\hline
\end{tabular}
\end{center}
\caption{FAT layer ablation experiments. Results are the values for the class accuracy on the ModelNet40 dataset.}
\label{spatialawarenessablationstable}
\end{table}

Certain combinations of local and global geometric properties work better than others in terms of performance. We combined the global and local vectors at different stages in the network to test our hypotheses. We do not use any attention mechanisms here. Our first experiment comprised of running two parallel blocks of embeddings, each looking at global (PointNet), and local (DGCNN) geometric properties respectively. Global feature aggregation over the outputs of these embeddings was concatenated to give a vector that comprised of information from both networks. This was then fed into a three-layer MLP for classification. We also computed, and then combined, the global and local feature vectors per layer, and computed further embeddings based on this combination.
The combination per layer results in a better accuracy than combining the features at the end.

We then added residual connections to this setting which further improved the performance of the networks by 0.4\% and we used this as our final architecture. Further experiments were conducted, such as removing the fully connected layers, but they all resulted in a drastic drop in the accuracy of the network. This leads us to believe that the fully connected layers are an important part of the architecture. We did try to increase the number of FAT layers to create a much deeper network that does not require fully connected layers and observed that the model's ability to learn point cloud processing tasks does improve with deeper layers. Unfortunately, this also considerably increases the training time and so we did not proceed further with these experiments. Our results are summarized in Table \ref{spatialawarenessablationstable}. The block diagrams to visualize these ablations are shown in 
Fig.~ \ref{spatialablations}.

\begin{figure}[h]
\begin{center}
\includegraphics[width=1.0\linewidth]{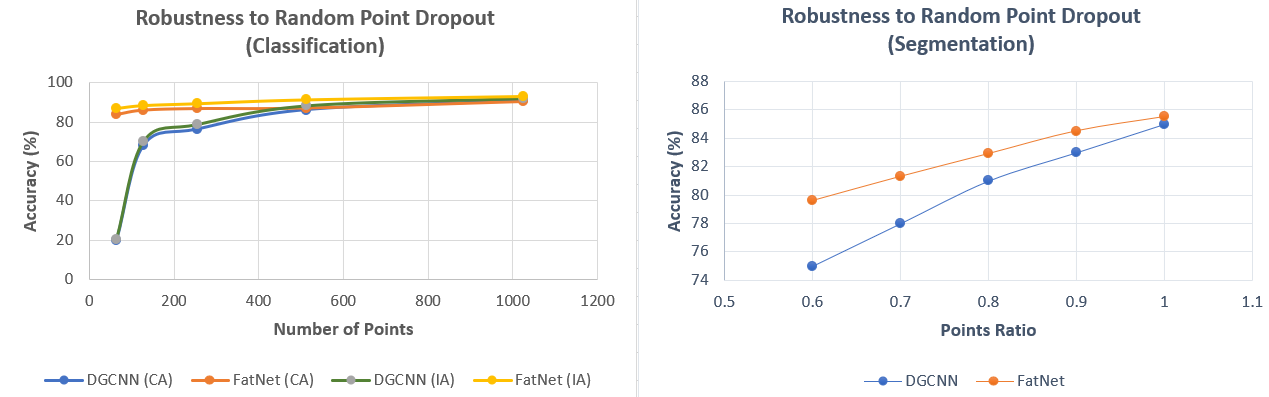}
\end{center}
   \caption{Robustness to sparse point inputs, compared with DGCNN.}
\label{Robust}
\end{figure}

\subsubsection{Robustness to missing points} 

To test how our network would do with sparse input points, we observed the degradation of our network's accuracy with respect to a small number of input points. We ran the network for 75 epochs with the same hyperparameters as discussed in Sect.~\ref{ch5:eval:classify}, and recorded the best test set accuracy. Surprisingly, the accuracy of FatNet  does not degrade to a large extent, even when the number of points input to the network drops to 128, showing how robust our network is to random point dropout. Figure \ref{Robust} (left) shows the comparison of our observation with DGCNN. The performance of DGCNN reduces considerably when the number of points drop below 384, while our network manages to maintain a 86.1\% accuracy, even with 128 points. 

\subsection{Ablation Studies}
\label{ablation}

We now shift our attention to show that the final architecture created, is in-fact the most optimal setting of the building blocks that it is comprised of. We single out four reasons why we believe that our architecture outperforms the current state of the art. These factors are: i) the FAT layers, ii) feature attention in the FAT layer, iii) skip connections between the layers and iv) the global feature aggregation unit. We have already shown the superiority of having feature-attentive layers in Table \ref{ModelNet40}, so we omit it for this experiment. Further, in Sect.~\ref{spatialawareness}, we have already shown how skip connections learn better feature representations. So we experiment with removing the feature attention block, while keeping the rest of the network as proposed, and then changing the global feature aggregation unit with max pooling based global aggregation, keeping the rest of the network as proposed, and report our results. In both cases, we observe a decrease in the performance of the architecture, showing that they are important components of our proposed architecture. The proposed global feature aggregation unit seems to play a large role in our network's performance, as it is responsible for almost a 0.4\% increase in accuracy. Table \ref{tab::ablation} shows the results of the ablation tests.

\begin{table}[h]
\begin{center}
\begin{tabular}{p{5.0cm} p{1.5cm}}
\hline
Transformation & Acc (\%) \\
\hline\hline
FatNet - Feature Attention in FAT Layer & 93.0\\
FatNet - Global Feature Aggregation Unit & 92.8\\
\hline
\end{tabular}
\end{center}
\caption{Summarising the two ablation settings experimented with.}
\label{tab::ablation}
\end{table}

We also experimented with different types of aggregation functions in our global feature aggregation (GFA) block. Our options were concatenating (CA), max pooling (MPA), and adding the attention-scaled pooling values (ours). Concatenating these values after attention scaling performed the worst. Max pooling provided good results, but addition gave the best results and we used this method in our final architecture. Table \ref{globalagg} summarises our results. \\

\begin{table}[h]
\begin{center}
\begin{tabular}{l c}
\hline
Method & Instance Accuracy (\%) \\
\hline\hline
PointNet + MP & 88.8\\
PointNet + CA & 89.1\\
PointNet + MPA & 89.2\\
PointNet + Ours & 89.4\\
FatNet + Ours & 93.2\\
\hline
\end{tabular}
\end{center}
\caption{Global feature aggregation ablations. PointNet results are based on our Keras re-implementation that achieves 88.8\% instance accuracy (Original instance accuracy - 89.2\%) Here, MP -max pooling, CA- concatenated attention, MPA - max pooling attention, Ours - FatNet  global feature aggregation block.}
\label{globalagg}
\end{table}

\section{Evaluation of Segmentation Performance}
\label{ch5:eval:segment}
We now compare the performance of our network with respect to the state of the art in 3D point cloud segmentation. For this section, the architecture is the same as the segmentation architecture shown in the bottom branch of 
Fig.~ \ref{final_arch}. We embed the transformer aligned 3D points into a higher dimensional space using FAT layers, and finally use our FatNet  global aggregation unit to get a point cloud statistic. Residual connections feed information to a final layer before the global aggregation unit, where it is concatenated and processed by a single FAT layer. We use our global aggregation statistic as the input into a three-layer MLP, which is then reshaped to give the point-wise segmentation result.

\subsection{3D Part Segmentation}

For this section, we use the ShapeNet part dataset \cite{shapenetpart}. Given a 3D point cloud, the task is to segment semantic parts of the point cloud. The dataset contains 16,881 3D models of 16 object categories with 50 part segmentation ground truths.
\begin{table}[h]
\begin{center}
\begin{tabular}{p{3.0cm} | p{1.5cm} }
\hline
Method & IoU \\
\hline\hline
Kd-Net \cite{kdnet} & 82.3  \\
SO-Net \cite{sonet} & 84.6  \\
RSNet \cite{RSNet} & 84.9 \\
3DmFVNet \cite{ben20173d} & 84.3 \\
SyncSpecCNN \cite{syncspeccnn} & 84.7 \\
PointNet++ \cite{pointnet2} & 85.1 \\
SpiderCNN \cite{spidercnn} & 85.3 \\
SPLATNet \cite{splatnet} & 85.4 \\
PointCNN \cite{pointcnn} & \textbf{86.1} \\
PCNN \cite{pcnn} & 85.1 \\
PointNet \cite{pointnet} & 83.7 \\
DGCNN \cite{dgcnn} & 85.1 \\
\hline
FatNet (MP) & 85.3  \\
FatNet & \textbf{85.5} \\
\hline
\end{tabular}
\end{center}
\caption{Part segmentation results on the ShapeNet part dataset.}
\label{part_segt}
\end{table}
Table \ref{part_segt} shows our results on the dataset. The evaluation metric for this task is the mean intersection over union (mIoU) for all the shapes in a particular category. It is computed by averaging out the IoUs of all the different shapes belonging to each object category. We use the official train/val/test split for consistency with other results.

We obtain two results for this task. The first result is obtained using FatNet  with max pooling (MP) and the other, with the full architecture, as shown in Fig.~\ref{final_arch}. Point cloud segmentation has been shown in PointNet to be sensitive to combining the right type of global and local features to obtain the best results. Hence, it is no surprise that our architecture shows competitive performance compared to existing architectures that use raw points as input in terms of the overall benchmark metric on the test set. We get an overall mIOU of 85.5\%, which is 0.4\% higher than that obtained by PointNet++ (which uses point and normals) and DGCNN, and 1.8\% higher than PointNet. Out of the 16 categories in the ShapeNet dataset, FatNet  gets the best results in six categories, which is the joint highest with DGCNN. Even on the objects that the architecture does not get the best in class results, it mostly gets a IoU value close to the best for that object.

\subsection{Robustness to random point dropout}

We observe the robustness of our segmentation architecture by dropping out points randomly and observing the effect on the network performance. We directly compare with the results presented in DGCNN. Our architecture is robust and accurate, providing a mIOU of almost 80\%, even when almost half the points are dropped from the point cloud. The performance of our architecture is far beter in general, compared to that of DGCNN in terms of robustness to fewer input points. The mIoU of DGCNN drops down steeply with a decrease in the number of points, while our architecture maintains a robust performance. The graph showing the quantitative differences between the performance of DGCNN and FatNet for this task can be visualized in Fig.~\ref{Robust} (right).

\section{Conclusion}
\label{ch5:summary}
We proposed FatNet, an architecture for 3D point cloud processing that employs feature attention and residual learning  to achieve state of the art results on benchmark point cloud analysis tasks. We hypothesised that feature-attentive layers can increase performance by incorporating additional relevance context. Our idea was validated
by the performance of FatNet on benchmark tasks and through a series of ablation tests. Our architecture is robust to random point dropout and provides a good trade-off between model complexity and performance. Our feature-attentive layer is a simple novel performance enhancement that can be easily incorporated into any existing 3D point cloud processing pipeline.

\bibliographystyle{IEEEtran}
\bibliography{IEEEexample}

\begin{thebibliography}{10}
\providecommand{\url}[1]{#1}
\csname url@samestyle\endcsname
\providecommand{\newblock}{\relax}
\providecommand{\bibinfo}[2]{#2}
\providecommand{\BIBentrySTDinterwordspacing}{\spaceskip=0pt\relax}
\providecommand{\BIBentryALTinterwordstretchfactor}{4}
\providecommand{\BIBentryALTinterwordspacing}{\spaceskip=\fontdimen2\font plus
\BIBentryALTinterwordstretchfactor\fontdimen3\font minus
  \fontdimen4\font\relax}
\providecommand{\BIBforeignlanguage}[2]{{%
\expandafter\ifx\csname l@#1\endcsname\relax
\typeout{** WARNING: IEEEtran.bst: No hyphenation pattern has been}%
\typeout{** loaded for the language `#1'. Using the pattern for}%
\typeout{** the default language instead.}%
\else
\language=\csname l@#1\endcsname
\fi
#2}}
\providecommand{\BIBdecl}{\relax}
\BIBdecl

\bibitem{pointnet}
C.~R. Qi, H.~Su, K.~Mo, and L.~J. Guibas, ``Pointnet: Deep learning on point
  sets for 3d classification and segmentation,'' in \emph{Proceedings of the
  IEEE conference on computer vision and pattern recognition}, 2017, pp.
  652--660.

\bibitem{deepsets}
\BIBentryALTinterwordspacing
M.~Zaheer, S.~Kottur, S.~Ravanbakhsh, B.~Poczos, R.~R. Salakhutdinov, and A.~J.
  Smola, ``Deep sets,'' in \emph{Advances in Neural Information Processing
  Systems 30}, I.~Guyon, U.~V. Luxburg, S.~Bengio, H.~Wallach, R.~Fergus,
  S.~Vishwanathan, and R.~Garnett, Eds.\hskip 1em plus 0.5em minus 0.4em\relax
  Curran Associates, Inc., 2017, pp. 3391--3401. [Online]. Available:
  \url{http://papers.nips.cc/paper/6931-deep-sets.pdf}
\BIBentrySTDinterwordspacing

\bibitem{pointnet2}
C.~R. Qi, L.~Yi, H.~Su, and L.~J. Guibas, ``Pointnet++: Deep hierarchical
  feature learning on point sets in a metric space,'' in \emph{Advances in
  neural information processing systems}, 2017, pp. 5099--5108.

\bibitem{dgcnn}
Y.~Wang, Y.~Sun, Z.~Liu, S.~E. Sarma, M.~M. Bronstein, and J.~M. Solomon,
  ``Dynamic graph cnn for learning on point clouds,'' \emph{Acm Transactions On
  Graphics (tog)}, vol.~38, no.~5, pp. 1--12, 2019.

\bibitem{se}
J.~Hu, L.~Shen, and G.~Sun, ``Squeeze-and-excitation networks,'' in
  \emph{Proceedings of the IEEE conference on computer vision and pattern
  recognition}, 2018, pp. 7132--7141.

\bibitem{Resnet}
K.~He, X.~Zhang, S.~Ren, and J.~Sun, ``Deep residual learning for image
  recognition,'' in \emph{Proceedings of the IEEE conference on computer vision
  and pattern recognition}, 2016, pp. 770--778.

\bibitem{mvcnn}
H.~Su, S.~Maji, E.~Kalogerakis, and E.~Learned-Miller, ``Multi-view
  convolutional neural networks for 3d shape recognition,'' in
  \emph{Proceedings of the IEEE international conference on computer vision},
  2015, pp. 945--953.

\bibitem{voxnet}
D.~Maturana and S.~Scherer, ``{VoxNet: A 3D Convolutional Neural Network for
  Real-Time Object Recognition},'' in \emph{{IROS}}, 2015.

\bibitem{spatialcontext}
F.~Engelmann, T.~Kontogianni, A.~Hermans, and B.~Leibe, ``Exploring spatial
  context for 3d semantic segmentation of point clouds,'' in \emph{Proceedings
  of the IEEE International Conference on Computer Vision Workshops}, 2017, pp.
  716--724.

\bibitem{frustrum}
C.~R. Qi, W.~Liu, C.~Wu, H.~Su, and L.~J. Guibas, ``Frustum pointnets for 3d
  object detection from rgb-d data,'' in \emph{Proceedings of the IEEE
  conference on computer vision and pattern recognition}, 2018, pp. 918--927.

\bibitem{pcpnet}
P.~Guerrero, Y.~Kleiman, M.~Ovsjanikov, and N.~J. Mitra, ``Pcpnet learning
  local shape properties from raw point clouds,'' in \emph{Computer Graphics
  Forum}, vol.~37, no.~2.\hskip 1em plus 0.5em minus 0.4em\relax Wiley Online
  Library, 2018, pp. 75--85.

\bibitem{sonet}
J.~Li, B.~M. Chen, and G.~Hee~Lee, ``So-net: Self-organizing network for point
  cloud analysis,'' in \emph{Proceedings of the IEEE conference on computer
  vision and pattern recognition}, 2018, pp. 9397--9406.

\bibitem{spidercnn}
Y.~Xu, T.~Fan, M.~Xu, L.~Zeng, and Y.~Qiao, ``Spidercnn: Deep learning on point
  sets with parameterized convolutional filters,'' in \emph{Proceedings of the
  European Conference on Computer Vision (ECCV)}, 2018, pp. 87--102.

\bibitem{pointcnn}
Y.~Li, R.~Bu, M.~Sun, W.~Wu, X.~Di, and B.~Chen, ``Pointcnn: Convolution on
  x-transformed points,'' in \emph{Advances in Neural Information Processing
  Systems 31}, S.~Bengio, H.~Wallach, H.~Larochelle, K.~Grauman,
  N.~Cesa-Bianchi, and R.~Garnett, Eds.\hskip 1em plus 0.5em minus 0.4em\relax
  Curran Associates, Inc., 2018, pp. 820--830.

\bibitem{seobjectdetection}
Y.~Cao, J.~Xu, S.~Lin, F.~Wei, and H.~Hu, ``Gcnet: Non-local networks meet
  squeeze-excitation networks and beyond,'' in \emph{Proceedings of the IEEE
  International Conference on Computer Vision Workshops}, 2019, pp. 0--0.

\bibitem{sesegmentation}
A.~G. Roy, N.~Navab, and C.~Wachinger, ``Concurrent spatial and channel squeeze
  and excitation in fully convolutional networks,'' 2018.

\bibitem{sesceneclass}
M.~{Juneja}, A.~{Vedaldi}, C.~V. {Jawahar}, and A.~{Zisserman}, ``Blocks that
  shout: Distinctive parts for scene classification,'' in \emph{2013 IEEE
  Conference on Computer Vision and Pattern Recognition}, June 2013, pp.
  923--930.

\bibitem{shapenet}
\BIBentryALTinterwordspacing
Z.~Wu, S.~Song, A.~Khosla, X.~Tang, and J.~Xiao, ``3d shapenets for 2.5d object
  recognition and next-best-view prediction,'' \emph{CoRR}, vol. abs/1406.5670,
  2014. [Online]. Available: \url{http://arxiv.org/abs/1406.5670}
\BIBentrySTDinterwordspacing

\bibitem{tensorflow}
\BIBentryALTinterwordspacing
M.~Abadi, A.~Agarwal, P.~Barham, E.~Brevdo, Z.~Chen, C.~Citro, G.~S. Corrado,
  A.~Davis, J.~Dean, M.~Devin, S.~Ghemawat, I.~Goodfellow, A.~Harp, G.~Irving,
  M.~Isard, Y.~Jia, R.~Jozefowicz, L.~Kaiser, M.~Kudlur, J.~Levenberg,
  D.~Man\'{e}, R.~Monga, S.~Moore, D.~Murray, C.~Olah, M.~Schuster, J.~Shlens,
  B.~Steiner, I.~Sutskever, K.~Talwar, P.~Tucker, V.~Vanhoucke, V.~Vasudevan,
  F.~Vi\'{e}gas, O.~Vinyals, P.~Warden, M.~Wattenberg, M.~Wicke, Y.~Yu, and
  X.~Zheng, ``{TensorFlow}: Large-scale machine learning on heterogeneous
  systems,'' 2015, software available from tensorflow.org. [Online]. Available:
  \url{https://www.tensorflow.org/}
\BIBentrySTDinterwordspacing

\bibitem{adam}
\BIBentryALTinterwordspacing
D.~P. Kingma and J.~Ba, ``Adam: {A} method for stochastic optimization,''
  \emph{CoRR}, vol. abs/1412.6980, 2014. [Online]. Available:
  \url{http://arxiv.org/abs/1412.6980}
\BIBentrySTDinterwordspacing

\bibitem{subvolume}
C.~R. Qi, H.~Su, M.~Nie{\ss}ner, A.~Dai, M.~Yan, and L.~J. Guibas, ``Volumetric
  and multi-view cnns for object classification on 3d data,'' in
  \emph{Proceedings of the IEEE conference on computer vision and pattern
  recognition}, 2016, pp. 5648--5656.

\bibitem{ECC}
M.~Simonovsky and N.~Komodakis, ``Dynamic edge-conditioned filters in
  convolutional neural networks on graphs,'' in \emph{Proceedings of the IEEE
  conference on computer vision and pattern recognition}, 2017, pp. 3693--3702.

\bibitem{kdnet}
R.~Klokov and V.~Lempitsky, ``Escape from cells: Deep kd-networks for the
  recognition of 3d point cloud models,'' in \emph{Proceedings of the IEEE
  International Conference on Computer Vision}, 2017, pp. 863--872.

\bibitem{pcnn}
\BIBentryALTinterwordspacing
M.~Atzmon, H.~Maron, and Y.~Lipman, ``Point convolutional neural networks by
  extension operators,'' \emph{ACM Trans. Graph.}, vol.~37, no.~4, pp.
  71:1--71:12, Jul. 2018. [Online]. Available:
  \url{http://doi.acm.org/10.1145/3197517.3201301}
\BIBentrySTDinterwordspacing

\bibitem{shapenetpart}
\BIBentryALTinterwordspacing
L.~Yi, V.~G. Kim, D.~Ceylan, I.-C. Shen, M.~Yan, H.~Su, C.~Lu, Q.~Huang,
  A.~Sheffer, and L.~Guibas, ``A scalable active framework for region
  annotation in 3d shape collections,'' \emph{ACM Trans. Graph.}, vol.~35,
  no.~6, pp. 210:1--210:12, Nov. 2016. [Online]. Available:
  \url{http://doi.acm.org/10.1145/2980179.2980238}
\BIBentrySTDinterwordspacing

\bibitem{RSNet}
Q.~Huang, W.~Wang, and U.~Neumann, ``Recurrent slice networks for 3d
  segmentation of point clouds,'' in \emph{Proceedings of the IEEE Conference
  on Computer Vision and Pattern Recognition}, 2018, pp. 2626--2635.

\bibitem{ben20173d}
Y.~Ben-Shabat, M.~Lindenbaum, and A.~Fischer, ``3d point cloud classification
  and segmentation using 3d modified fisher vector representation for
  convolutional neural networks,'' \emph{arXiv preprint arXiv:1711.08241},
  2017.

\bibitem{syncspeccnn}
L.~Yi, H.~Su, X.~Guo, and L.~J. Guibas, ``Syncspeccnn: Synchronized spectral
  cnn for 3d shape segmentation,'' in \emph{Proceedings of the IEEE Conference
  on Computer Vision and Pattern Recognition}, 2017, pp. 2282--2290.

\bibitem{splatnet}
H.~Su, V.~Jampani, D.~Sun, S.~Maji, E.~Kalogerakis, M.-H. Yang, and J.~Kautz,
  ``Splatnet: Sparse lattice networks for point cloud processing,'' in
  \emph{Proceedings of the IEEE Conference on Computer Vision and Pattern
  Recognition}, 2018, pp. 2530--2539.

\end{thebibliography}

\end{document}